\documentclass{article}
\usepackage{spconf,amsmath,graphicx}
\usepackage{xcolor}
\usepackage{mathrsfs}

\usepackage{mathtools}
\DeclarePairedDelimiter\abs{\lvert}{\rvert}%
\DeclarePairedDelimiter\norm{\lVert}{\rVert}%
\makeatletter
\let\oldabs\abs
\def\abs{\@ifstar{\oldabs}{\oldabs*}}
\let\oldnorm\norm
\def\norm{\@ifstar{\oldnorm}{\oldnorm*}}
\makeatother

\renewcommand{\bar}{\overline}
\usepackage[normalem]{ulem} 

\usepackage{subfig}

\def\x{{\mathbf x}}

\newcommand{\ubar}[1]{\text{\b{$#1$}}}

\title{Evaluating Crowd Density Estimators via Their Uncertainty Bounds}
\name{Jennifer Vandoni, Emanuel Aldea and Sylvie Le H\'{e}garat-Mascle\thanks{This work was funded by French ANR grant ANR-15-CE39-0005.}}
\address{SATIE - CNRS UMR 8029, Paris-Sud University, Paris-Saclay University, France}
\begin{document}
%
\maketitle
\begin{abstract}
In this work, we use the Belief Function Theory which extends the probabilistic framework in order to provide uncertainty bounds to different categories of crowd density estimators. Our method allows us to compare the multi-scale performance of the estimators, and also to characterize their reliability for crowd monitoring applications requiring varying degrees of prudence.  
\end{abstract}
%

\section{Introduction}
\label{sec:intro}

Understanding crowd systems and predicting their evolution is of paramount importance when considering the world population growth and urbanization rates. One of the reasons for which the urban infrastructure sector has not fully taken advantage of vast available video data is the difficulty to extract accurately microscopic and macroscopic observations in high-density conditions. Although it does not require accurate target localization, density estimation inside crowds is still a challenging problem, due to phenomena such as strong occlusion and visual homogeneity. However, deep learning advancements significantly improved the state-of-the-art performance (see~\cite{sindagi2018survey} for a comprehensive survey).
Recent methods are mostly based on the estimation of a density map whose integral over a region provides the number of people within it, in such a way to incorporate spatial information directly into the learning process
(e.g. MCNN~\cite{zhang2016single}, Cascaded-MTL~\cite{sindagi2017cnn}, Hydra CNN~\cite{onoro2016towards}, CSRNet~\cite{li2018csrnet}). In the vast majority of the proposed works, the estimator evaluation is performed at image scale, with error metrics such as MAE or MSE~\cite{sindagi2018survey}. However, from the point of view of modelling the crowd as a dynamical system, accurate \textit{local} densities are required in order to characterize wave-like propagation phenomena. The error metrics above are related to large scale statistics, which do not apply to small scales due to compensation between overestimating and underestimating the density in different areas. 
An additional limitation of current density estimators is the absence of an uncertainty range provided along with the scalar density. Ranges on the pedestrian count are greatly needed, as the trade-off between safety concerns and optimal use of infrastructure capacity promotes  different levels of congestion in different contexts. In statistical learning theory, the uncertainty of classification or regression processes has been studied based on how models are affected by the skewed distribution and noise of training data, by inaccurate training data labeling or by the regularization policy. SVM output uncertainty has been studied~\cite{xu2016evidential} due to algorithm's convenient generalization ability coupled to the simple underlying principle. In \cite{gal2016dropout} a new strategy was introduced for addressing the epistemic uncertainty estimation in Deep Neural Networks by approximating the probabilistic output distribution using dropout during inference. 

In this line of research, we propose a generic approach for evaluating the uncertainty of the output of a crowd density estimator. As a second contribution, we apply the proposed evaluation on a multiscale domain derived from the image lattice, which allows us to characterize the estimator performance locally as well. We show that we are able to compare different learning algorithms across the scale space and to provide density estimations with bounded uncertainties. 

\section{Evidential CNN-ensemble }
\label{sec:cnn_ensemble}
\noindent \textbf{FE+LFE network.} Following recent advancements on density estimation~\cite{li2018csrnet}, we propose a fully convolutional network which makes use of dilated convolutions instead of pooling layers, in order to preserve the output resolution in presence of small targets. However, as highlighted in~\cite{hamaguchi2018effective}, aggressively increasing dilation factors through the network layers is detrimental in aggregating local features. By taking inspiration from this latter work we propose a network which is composed of two parts, i.e. a front end (FE) module with increasing dilation factors to consider larger context around small objects, and a local feature extractor (LFE) module with decreasing dilation factors to enforce the spatial consistency of the output by gathering spatial information. Moreover, unlike~\cite{hamaguchi2018effective}, we add batch normalization before ReLU activation functions for faster convergence. The structure of the proposed FE+LFE network is detailed in Table~\ref{tab:lfe}. The number of filters per layer is kept small to avoid overfitting since we intend to be able to train the network with relatively small datasets. Note that we employ a ReLU activation function also after the last layer. This has the effect of a zero-threshold; nevertheless, 
it has beneficial effects on backpropagation with respect to a simple post-processing thresholding. 
The local density estimation is therefore enhanced, since the network loses its tendency to add noise to compensate between low and high values.

Finally, a L2 loss function is used between the estimated density map and the ground-truth derived by placing a Gaussian on each head center as in~\cite{zhang2016single}. Since we know the geometry of the scene, we apply perspective correction as in~\cite{vandoni2017active} instead of geometry-adaptive kernels.

\begin{table}[t!]
{
\scriptsize
\centering
\begin{minipage}{.2\textwidth}
\centering
\begin{tabular}{|c|c|}
\hline
 & Layers - part 1\\
\hline
FE &  Conv $3 \times 3$, $F=16$, $D=1$  \\
          &  Conv $3 \times 3$, $F=32$, $D=1$  \\
          &  Conv $3 \times 3$, $F=32$, $D=2$  \\
          &  Conv $3 \times 3$, $F=64$, $D=2$  \\
          \hline
          &  Conv $3 \times 3$, $F=64$, $D=3$  \\
          \hline
\end{tabular}
\end{minipage} \;\;\;\;\;\;\;\;
\begin{minipage}{.2\textwidth}
\centering
\begin{tabular}{|c|c|}
\hline
 & Layers - part 2\\
\hline
LFE       &  Conv $3 \times 3$, $F=64$, $D=2$  \\
          &  Conv $3 \times 3$, $F=64$, $D=2$  \\
          &  Conv $3 \times 3$, $F=64$, $D=1$  \\
          &  Conv $3 \times 3$, $F=64$, $D=1$  \\
          \hline
          &  Conv $1 \times 1$, $F=1$, $\;\:D=1$  \\
          \hline
\end{tabular}
\end{minipage}
}

\caption{Architecture of the FE+LFE network.
$F$ is the number of filters and $D$ is the dilation factor of dilated convolutions.
Each convolutional layer is followed by batch normalization (except for the last one) and ReLU activation function. }
\label{tab:lfe}
\end{table}%
%
%
%
%

\noindent \textbf{Building a CNN-ensemble.} Recently ensemble techniques have been successfully exploited by the deep learning community since they allows for more robust predictions as well as for a measure of predictive uncertainty, i.e. the confidence of the network with respect to its prediction (which in our case represents the likelihood of head presence). 
In the context of Bayesian Neural Networks (BNNs), the authors of~\cite{gal2016dropout} developed a new theoretical framework called \textit{MC-dropout} casting dropout~\cite{srivastava2014dropout} as approximate Bayesian inference in Gaussian processes. This method overcomes the major limitations of BNNs that generally require prohibitive computational costs~\cite{neal2012bayesian,kingma2013auto, blundell2015weight}.
In~\cite{gal2016dropout} instead, after training the network, Monte Carlo (MC) methods are used at inference time to draw samples from a Bernoulli distribution across the network weights, by performing \textit{T} stochastic forward passes through the network with dropout. The ensemble is thus composed by \textit{T} different realizations given by dropping out different units of the network at each forward pass. 
Another ensemble approach (although non-Bayesian) has been recently proposed in~\cite{lakshminarayanan2017simple}, where a \textit{deep ensemble} is derived by training the same network on the same data but with different random weight initializations.
Compared to MC-dropout, this method has nonetheless the immediate drawback of requiring multiple training of the network. 

In this work, we derive a CNN-ensemble relying on MC-dropout. We train the network once and then we sample the posterior distribution over the weights using dropout at inference time, obtaining \textit{T} different realization maps $\mathscr{\hat{M}}_1,\dots,\mathscr{\hat{M}}_T$, outputs of different dropout-perturbed versions of the original network. Classically, the mean map $\mathscr{M}_\mu$, given by the mean value evaluated independently for each pixel, would be interpreted as the final prediction map, while the standard deviation map $\mathscr{M}_\sigma$ would be interpreted as an estimate of the predictive uncertainty. However, we propose to work in the Belief Function (BF) framework~\cite{Shafer1976,Smets1994}, that we consider more suited to model the specific imprecision of each different realization obtained with dropout, allowing us to derive the uncertainty bounds for density estimation.

\noindent \textbf{Modeling imprecision with BFT.} To handle both the uncertainty provided by the classification and the related imprecision that may exist due to the specific classifier and/or data used in the training process, 
Belief Function Theory (BFT), also called evidential theory, is designed to handle a larger hypothesis set than the probabilistic one. Denoting by $\Theta$ the discernment frame, i.e. the set of mutually exclusive hypotheses of cardinality $\left|\Theta\right|$, belief functions are defined on the powerset
$2^\Theta$.
In our setting, denoting by $H$ and $\bar{H}$ the two mutually exclusive (\textit{singleton}) hypotheses \textit{``Head''} and \textit{``Not~Head''}, the discernment frame is $\Theta=\left\{H,\bar{H}\right\}$ while $2^\Theta=\left\{\emptyset,H,\bar{H},\left\{H,\bar{H}\right\}\right\}$.
Classically, the \textit{mass} function $m$ is the \textit{Basic Belief Assignment} (BBA) that satisfies $\forall A\in 2^\Theta, \; m(A)\in \left[0,1\right]$, $\sum_{A\in2^\Theta} m(A)=1$. The hypotheses associated to non-null mass functions are called \textit{focal elements}. BBAs that have only singleton hypotheses as focal elements are called Bayesian BBAs.

Now, we want to consider the imprecision that possibly arises performing inference on unknown images with a model learned by a neural network, by modelling the pixel-wise classification outputs as BBAs. 
We therefore exploit the the CNN-ensemble composed by the \textit{T} realizations obtained with MC-dropout.
Firstly, we derive Bayesian BBA maps $\mathcal{M}^\mathcal{B}_1,\dots,\mathcal{M}^\mathcal{B}_T$, where a BBA is associated to each pixel $\mathbf{x}$ of every realization, so that we obtain \textit{T} maps of BBAs  $\left\{m^{\mathcal{B}}_{\mathbf{x},t}\right\}_{\mathbf{x}\in\mathcal{P}}$, where $\mathcal{P}$ is the pixel domain and $t \in \left\{ 1,\dots,T\right\}$.  These Bayesian BBA maps are 4-layer images where each layer corresponds to the mass value of any hypothesis in $\left\{\emptyset,H,\bar{H},\Theta\right\}$ respectively. For example, $\mathcal{M}^\mathcal{B}_t(A)$ corresponds to the layer image associated to hypothesis $A$ for the realization (source) $t$. In this preliminary Bayesian BBA allocation, layer images corresponding to non-singleton hypotheses are null by definition, whereas for each source $t$, with $t=1,\dots,T$: 
$\mathcal{M}_{t}^\mathcal{B}(H)= \mathscr{\hat{M}}_t$, and $\mathcal{M}_{t}^\mathcal{B}(\bar{H})= 1 - \mathscr{\hat{M}}_t$.


In order to account for the reliability of the pixel-wise prediction given by every source, we perform a pixel-wise tailored discounting, namely a 
\textit{generalization}
of each BBA on the basis of its reliability~\cite{Shafer1976}.  
To evaluate this latter, noting that the median has been shown to be a more robust estimator than the average in presence of outliers, for each source $t$ we compute a discounting coefficient map $\Gamma_t \;: \; \left\{ \gamma_{\mathbf{x},t} \right\}_{\mathbf{x}\in\mathcal{P}}$ such that a different coefficient $\gamma_{\mathbf{x},t}$ is associated to every pixel of each source,
\begin{equation}
\label{eq:discounting}
    \Gamma_t = \alpha \left ( 1 - \left( \abs{ \mathscr{\hat{M}}_t - \text{median} \left( \left\{ \mathscr{\hat{M}} \right\}_1^T \right)  } \right) \right).
\end{equation}
In this way, we discount more pixels whose value is more distant to the median value among the \textit{T} realizations, since they are supposed to be less representative (even possibly outliers). The $\alpha$ parameter is a scaling factor which allows us to control the amount of discounting.
Applying the proposed discounting, we derive the following BBAs map for every source $t$: $\forall A\in\left\{H,\bar{H}\right\}$,
\begin{eqnarray}
\label{eq:cnn_bba}
 \left\{
 \begin{array}{rcl}
\mathcal{M}_{t}(\emptyset)&=&\left\{0\right\}_{\mathbf{x}\in\mathcal{P}}, \\
 \mathcal{M}_{t}(A)&=& \Gamma_{t} \star \mathcal{M}_{t}^\mathcal{B}(A), \\
\mathcal{M}_{t}(\Theta)& =& \left\{1\right\}_{\mathbf{x}\in\mathcal{P}}-\mathcal{M}_{t}(H)-\mathcal{M}_{t}(\bar{H}), \\
\end{array}
\right.
\end{eqnarray}
where $M_1 \star M_2$ represents the Hadamard product between matrices $M_1$ and $M_2$.

To combine the \textit{T} different maps to obtain a single output map $\mathcal{M}$ with BBAs associated to each pixel $\mathbf{x}$, i.e. $\left\{m_{\mathbf{x}} \right\}_{\mathbf{x}\in\mathcal{P}}$, we use the conjunctive combination rule~\cite{Smets1994}. In our case where $\left|\Theta\right|=2$, the analytic result may be easily derived: $\forall A\in\left\{H,\bar{H}\right\}$, 
\begin{eqnarray}
\label{eq:conj_rule_analytic}
 \left\{
 \begin{array}{rcl}
m_{\mathbf{x}}\left(A\right) &=& 
\sum\limits_{\substack{\left(B_1,\dots,B_T\right) \in\left\{A,\Theta\right\}^T,\\ \exists t\in\left[1,T\right] s.t. B_t= A}} \prod_{t=1}^T m_{\mathbf{x},t}\left(B_t\right), \\
m_{\mathbf{x}}\left(\Theta\right) &=&
\prod_{t=1}^T m_{\mathbf{x},t}\left(\Theta\right),\\
m_{\mathbf{x}}\left(\emptyset\right) &=& 
1-m_{\mathbf{x}}\left(H\right)-m_{\mathbf{x}}\left(\bar{H}\right)-m_{\mathbf{x}}\left(\Theta\right).
\end{array}
\right.
\end{eqnarray}
The result is thus a four-layer map $\mathcal{M}$ of BBAs $m_\mathbf{x}$, that can be used to derive evidential measures of uncertainty about  the network prediction.
To this extent, we can obtain the ignorance map as $\mathcal{M}(\Theta)$, that represents the remaining ignorance which has been decreased by the combination but not completely solved, indicating a lack of sufficient information during training to perform a reliable prediction. Likewise, $\mathcal{M}(\emptyset)$ is often interpreted as a conflict map~\cite{lachaize2018evidential}, and presents higher values for pixels whose prediction completely disagrees through the various realizations.

Finally, in every pixel $\mathbf{x}$ the decision is taken from $m_{\mathbf{x}}$. Pignistic probability~\cite{Smets1994} may be used to give a probabilistic interpretation to the BBAs. Since in our setting $\left|\Theta\right|=2$, $\forall A\in \left\{ H,\bar{H} \right\}$, 
$BetP_{\mathbf{x}}(A)=\frac{1}{1-m_{\mathbf{x}}(\emptyset)}  \left( m_{\mathbf{x}}(A)+\frac{m_{\mathbf{x}}(\Theta)}{2}\right)
$. This allows us to assign a $BetP_{\mathbf{x}}(H)$ value to the resulting BBA associated to each pixel $\mathbf{x}$ that will be differently normalized on the basis of its conflict value, $m_{\mathbf{x}}(\emptyset)$.

Then, other functions are in a one-to-one relationship with $m_\mathbf{x}$, and can be used either for decision or for some computations, namely the \textit{Plausibility} (\textit{Pl}) and the \textit{Belief} (\textit{Bel}) functions. In this particular setting where $\left|\Theta\right|=2$ applying a normalization to the BBAs (so that $m_{\mathbf{x}}(\emptyset)=0$), they are defined as: $Bel_{\mathbf{x}}(A)=\frac{1}{1-m_{\mathbf{x}}(\emptyset)}  \left( m_{\mathbf{x}}(A)\right)$, and $Pl_{\mathbf{x}}(A)=\frac{1}{1-m_{\mathbf{x}}(\emptyset)} \left( m_{\mathbf{x}}(A)+m_{\mathbf{x}}(\Theta) \right)$. These functions may also be interpreted as upper and lower probabilities respectively~\cite{Shafer1976} and they check the duality property: $\forall A\in 2^\Theta, Pl_{\mathbf{x}}(A)=1-Bel_{\mathbf{x}}(\bar{A})$ (where $\bar{A}$ represents the complement of $A$ with respect to $\Theta$).
\section{Density uncertainty for bounding pedestrian counts}
\label{sec:density_uncertainty}

In this work we propose a multiscale  evaluation strategy which computes for each considered scale $\mathcal{S}$  indicators based on all squared subdomains $S \in \mathcal{S}_i$. These indicators use the derived upper and lower density bounds $\ubar{s}(S)$, $\bar{s}(S)$: $\ubar{s}(S) = w \sum_{\x \in S} Bel_{\x}(H)$ and $\bar{s}(S) = w \sum_{\x \in S} Pl_{\x}(H) $. 
The factor $w$ relating the numerical output to the actual pedestrian count is 1 for networks trained on actual density maps, but in the general case it may be determined as in~\cite{lempitsky2010learning} on a validation set consisting of $BetP(H)$ maps. We then calculate for $\mathcal{S}_i$ the \textit{prediction error probability} (PEP) as:
\begin{equation} 
\label{eq:pepeq}
{PEP}_i = \Big| \{S \in \mathcal{S}_i | g(S) \in [\ubar{s}(S),\bar{s}(S)] \} \Big| / | \mathcal{S}_i| ,
\end{equation}
and the \textit{relative imprecision} (RI) interval as:
\begin{equation} 
\label{eq:rieq}
{RI}_i = \Big(\sum_{S \in \mathcal{S}_i}(\bar{s}(S)-\ubar{s}(S))/g(S)\Big) / | \mathcal{S}_i| ,
\end{equation}
where $g(S)$ is the ground-truth count over $S$. In our work, we take $\mathcal{S}_1$ as the set of the largest possible squares which fit the image space, and then we use a scale factor $\delta$ to reduce the square side for subsequent scales.

The RI criterion highlights the size of the imprecision interval around the estimated count, while the PEP criterion indicates the error rate of the prediction, namely whether the ground-truth count for the considered region is outside the estimated interval. Thus, a two-axis plot presenting the evolution of RI vs. PEP across multiple scales and for different estimators allows one to compare them and to select an operating point with an explicit uncertainty tied to a desired error rate. In order to compute the values required by Eqs.~\eqref{eq:pepeq} and~\eqref{eq:rieq}, the process may be accelerated significantly by using the Integral Histogram~\cite{porikli2005integral} trick, given that the most intensive task is to compute sums over rectangular supports defined in the bounded image space.

\section{Experimental results}
\label{sec:results}
We validated our proposed approach on high-density crowd images acquired at Makkah during Hajj~\cite{vandoni2019evidential}. Besides evaluating the proposed FE+LFE network, we compared it to U-Net~\cite{ronneberger2015u}, originally introduced for medical image segmentation and very effective even on relatively small training datasets as in our case (35 crowd images). The two networks are trained by using Adam stochastic optimizer with a learning rate of $7\times10^{-3}$ (FE+LFE) and of $10^{-2}$ (U-Net). 
Additionally, we perform data augmentation and early stopping in order to limit overfitting. A CNN-ensemble of size $T=10$ is then obtained by applying dropout at inference time in the central layers as in~\cite{kendall2015bayesian} with probability $p_{\text{drop}}=0.5$. 
\begin{figure}[htb]
\captionsetup[subfloat]{farskip=0pt,captionskip=0.1pt}
  \centering
  \subfloat[FE+LFE]{\includegraphics[height=3.5cm]{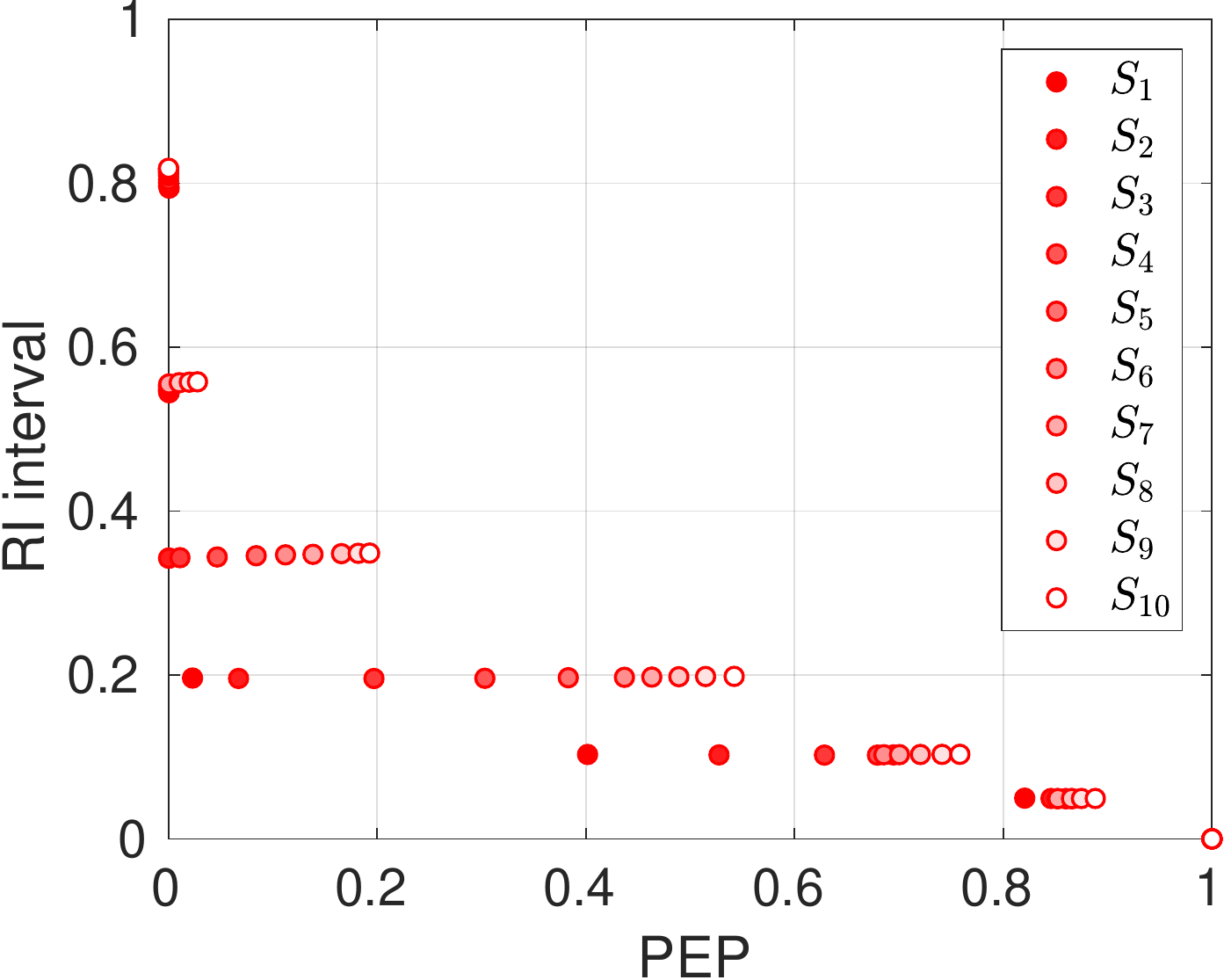}\label{fig:lfe1}}
  \centering
  \subfloat[U-Net~\cite{ronneberger2015u}]{\includegraphics[height=3.5cm]{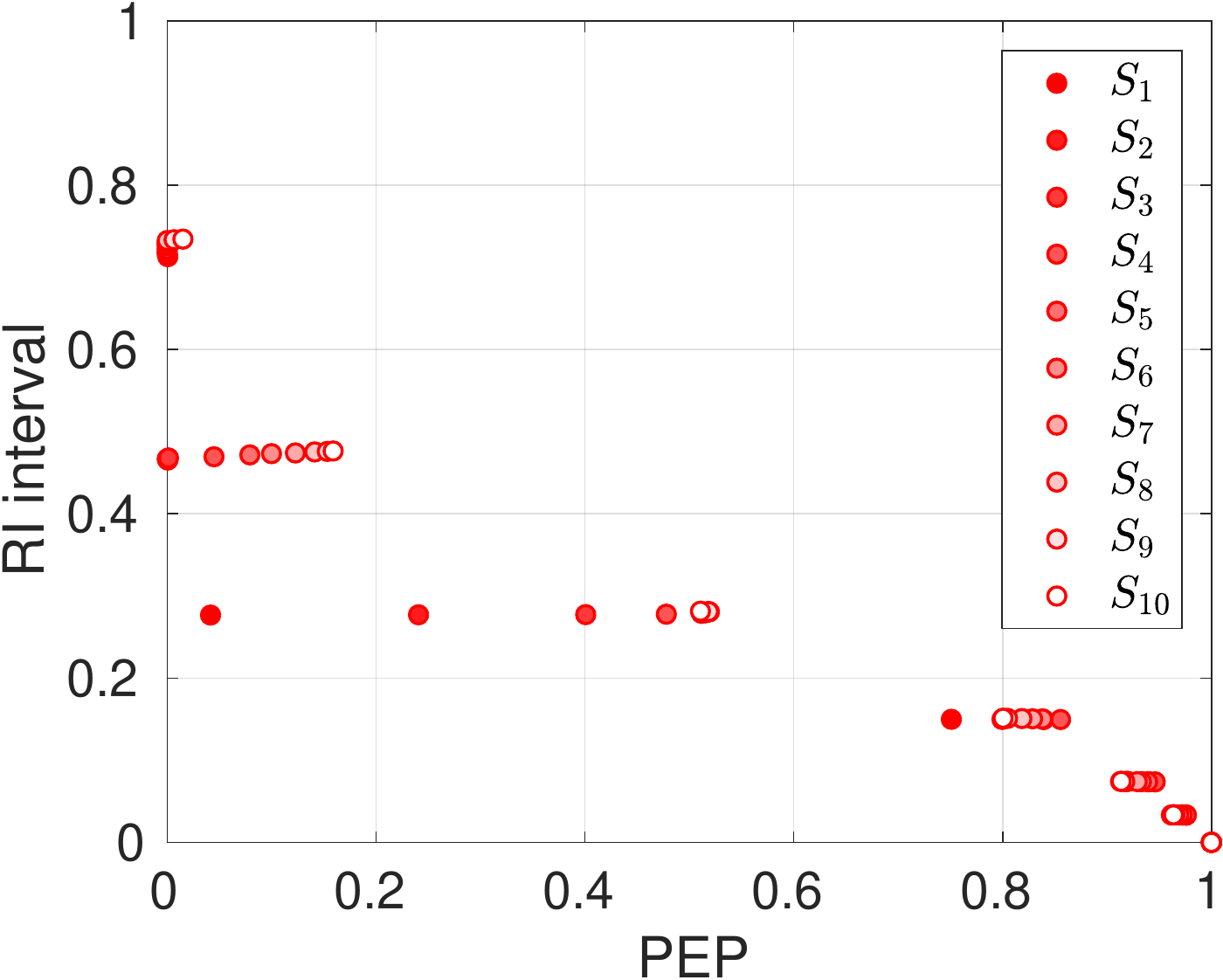}\label{fig:unet}} \\
%
  \centering
  \subfloat[FE+LFE (trained on less data)]{\includegraphics[height=3.5cm]{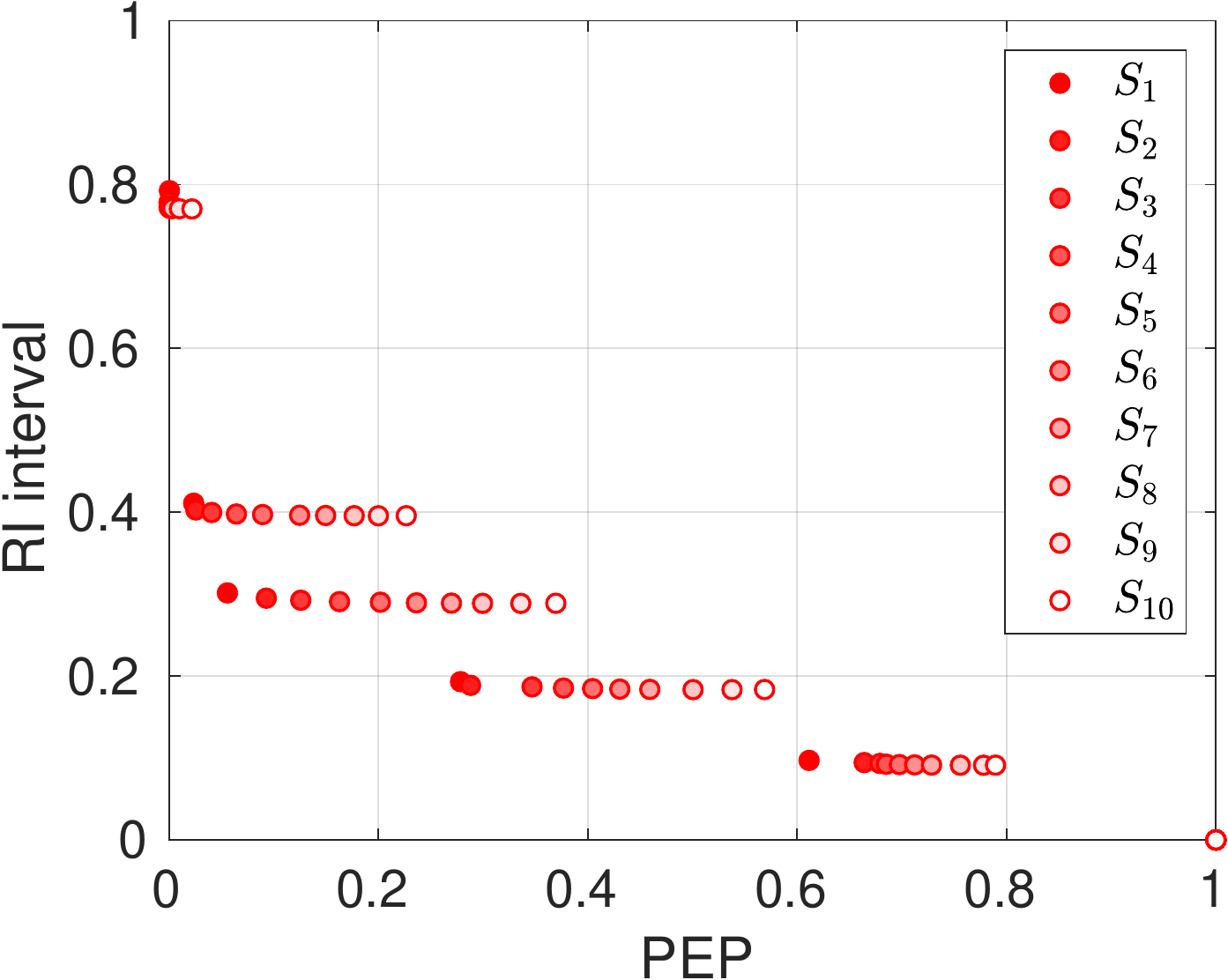}\label{fig:lfe2}}
  \centering
  \subfloat[SVM~\cite{vandoni2019evidential}]{\includegraphics[height=3.5cm]{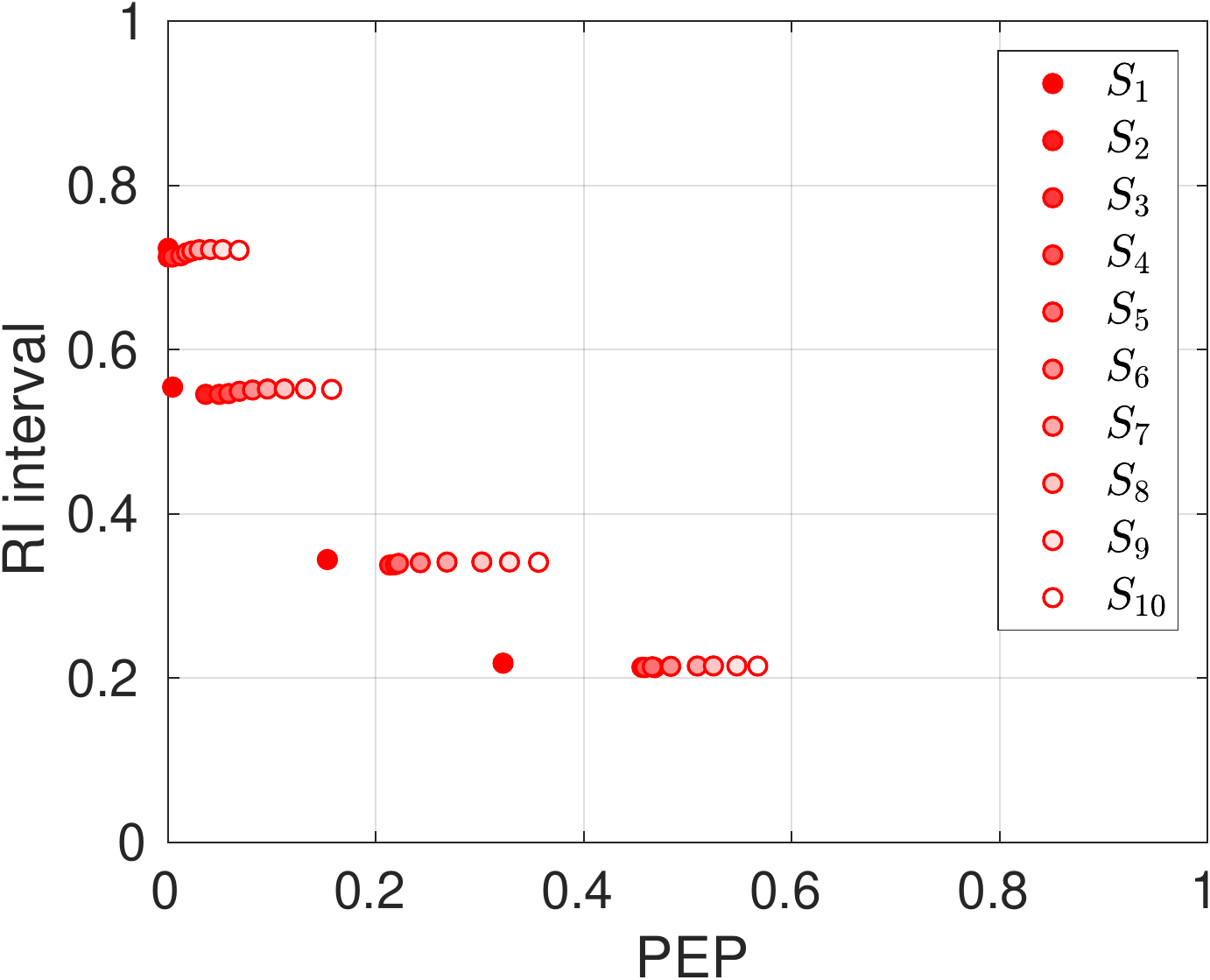}\label{fig:svm}} 
%
\caption{Density estimator evaluation with the proposed RI vs. PEP plot at multiple scales and with different discounting amounts. Each horizontal cluster corresponds to a different discounting factor.
\label{fig:res}}
\end{figure}

Figures~\ref{fig:lfe1} and~\ref{fig:unet} show the results when applying the proposed uncertainty bound evaluation for the FE+LFE and U-Net networks respectively. Ideally, an estimator should predict with a high confidence (low PEP) that the estimated count is within a small RI interval. One may increase the size of the RI interval by decreasing the $\alpha$ parameter in Eq.~\eqref{eq:discounting}, in order to obtain better prediction accuracy (at the expense of a larger RI). We tested different discounting factors, corresponding to horizontally aligned clusters of dots. For each cluster, each dot depicts the performance obtained at a different scale, with a scale factor $\delta=1.1$, $\mathcal{S}_1$ being the largest scale. 
Both networks perform better at larger scales, due to error compensation. The proposed FE+LFE network outperforms U-Net, showing the importance of preserving spatial information without pooling operations in presence of small targets, while increasing at the same time the contextual information with dilations.

To stress the independence of the proposed evaluation approach with respect to the classifier used, Fig.~\ref{fig:svm} shows the results of the density estimation obtained with SVM using active learning (AL) as in~\cite{vandoni2019evidential}, where an SVM-ensemble is built iteratively by training SVMs with different descriptors on selected informative samples. The imprecision derives both from possible errors in the calibration procedure to obtain probability estimates out of SVM scores, and from the score heterogeneity in the image space. Moreover, Fig.~\ref{fig:lfe2} shows the results obtained training the proposed FE+LFE network with a smaller amount of data (i.e. the pool of unlabeled samples $\mathcal{U}$ available for AL in~\cite{vandoni2019evidential}). This allows us to perform two different types of analysis. 
Firstly, we can perform a fairer comparison between the two classifiers.
To this extent, we notice that FE+LFE, even when trained on less data, outperforms the SVM-based approach, especially at larger scales. Nonetheless, the two methods exhibit almost identical performance when considering the smaller scales. 
Secondly, it is interesting to evaluate the same network trained with different amounts of data. According to Figs.~\ref{fig:lfe1} and~\ref{fig:lfe2}, we see that a larger training set is beneficial for density estimation especially at larger scales. However, considering smaller scales, the performance gap is consistently reduced, indicating thus an implicit limit in the network capacity (increasing the number of layers and/or filters per layer could help, paying attention to overfitting).
 
\begin{figure}[t!]
\captionsetup[subfloat]{farskip=0pt,captionskip=0.1pt}
  \centering
  \subfloat[Image patch $S$, $g(S)=12.3$ ]{\includegraphics[height=2.5cm]{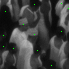}\label{fig:gtheads}} \;
  \centering
  \subfloat[$BetP(H)$ map, $s(S)=12.01$]{\includegraphics[height=2.5cm]{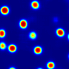}\label{fig:mapbetp}} \;
%
  \centering
  \subfloat[$\mathcal{M}(\Theta)$ map, $\bar{s}(S)-\ubar{s}(S)=3.2$]{\includegraphics[height=2.5cm]{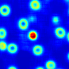}\label{fig:mapigno}}
%
\caption{Visual results of the density estimation map along with the estimated uncertainty bounds.
\label{fig:visualres}}
\end{figure}

Figure~\ref{fig:gtheads} shows an image patch with corresponding ground-truth count (obtained after Gaussian smoothing). 
Figure~\ref{fig:mapbetp} shows the resulting $BetP(H)$ map which represents the scalar density estimation map, while Fig.~\ref{fig:mapigno} shows the imprecision map $\mathcal{M}(\Theta)$ 
(in our case for pixel $\x$ the imprecision value $Pl_{\x}(H)-Bel_{\x}(H)$ is equal to $m_{\x}(\Theta)$).
The values in $\mathcal{M}(\Theta)$ may be interpreted as the
predictive uncertainty, and provide a bound for the density estimation itself.
For the given region $S$ indeed, by integrating over the $BetP(H)$ map we obtain the estimated number of people within it. Similarly, integrating over the $\mathcal{M}(\Theta)$ map we obtain the imprecision interval $\bar{s}(S)-\ubar{s}(S)$. Then, the corresponding RI interval is given by  $(\bar{s}(S)-\ubar{s}(S))/g(S)=0.26$, so that we can conclude that in $S$ there are $12.01 \pm 13\%$ heads, i.e. $s(S) \in [10.4, 13.6]$. Moreover, from Fig.~\ref{fig:mapigno} we can notice that, in addition to head edges, ignorance is particularly high on heads with lower gradient on the borders and strong clutter, reflecting in a smaller confidence about the prediction. Finally as expected, we underline the desirable effect of ignorance being higher in circularly-shaped areas (e.g. shoulders, or round dark blobs) which are similar to heads, even if they have a low corresponding score.





\section{Conclusion}
We proposed a strategy for associating an uncertainty interval to crowd density estimation using BFT. A new evaluation method taking into account the output uncertainty
at multiple scales was proposed as well. The results show that our contributions are effective in characterizing the multi-scale performance of different density estimators. Our work opens a promising avenue for crowd safety applications which account for estimation uncertainty during planning and monitoring. Future work will be devoted to applying our evaluation to other widely used density estimation networks such as MCNN or CSRNet across  more datasets.

\bibliographystyle{IEEEbib}
\bibliography{refs}

\end{document}